\pdfoutput=1
\documentclass[11pt]{article}
\usepackage[final]{acl}
\usepackage{times}
\usepackage{latexsym}

\usepackage[T1]{fontenc}
\usepackage[utf8]{inputenc}

\usepackage{microtype}

\usepackage{inconsolata}

\usepackage{graphicx}
\usepackage{multirow}
\usepackage{hyperref}
\usepackage{tcolorbox}
\title{EquivPruner: Boosting Efficiency and Quality in LLM-Based Search via Action Pruning}
\author{
 \textbf{Jiawei Liu\textsuperscript{1,2,3}}\thanks{Work done during an internship at iFLYTEK Research.},
 \textbf{Qisi Chen\textsuperscript{1}},
 \textbf{Jianshu Zhang\textsuperscript{3}},
 \textbf{Quan Liu\textsuperscript{3}},
 \textbf{Defu Lian\textsuperscript{1}}\thanks{$\quad$ Corresponding Author.},
\\
\\
 \textsuperscript{1}School of Computer Science and Technology, University of Science and Technology of China\\
 \textsuperscript{2}Department of Data Science, City University of Hong Kong\\
 \textsuperscript{3}iFLYTEK Research
\\
\texttt{ljw1222@mail.ustc.edu.cn}, 
\texttt{liandefu@ustc.edu.cn} \\
}

\begin{document}
\maketitle
\begin{abstract}
Large Language Models (LLMs) excel at complex reasoning through search algorithms, yet current strategies often suffer from massive token consumption due to redundant exploration of semantically equivalent steps. Existing semantic similarity methods struggle to accurately identify such equivalence in domain-specific contexts like mathematical reasoning. To address this, we propose \textit{EquivPruner}, a simple yet effective approach that identifies and prunes semantically equivalent actions during LLM reasoning search.  We also introduce MathEquiv, the first dataset we created for mathematical statement equivalence, which enables the training of a lightweight equivalence detector. Extensive experiments across various models and tasks demonstrate that \textit{EquivPruner} significantly reduces token consumption, improving searching efficiency and often bolstering reasoning accuracy. For instance, when applied to Qwen2.5-Math-7B-Instruct on GSM8K, \textit{EquivPruner} reduced token consumption by 48.1\% while also improving accuracy. Our code is available at \url{https://github.com/Lolo1222/EquivPruner}.
\end{abstract}

\section{Introduction}
Large Language Models (LLMs) are increasingly demonstrating remarkable capabilities, yet their performance scaling during pretraining faces potential constraints due to data limitations \citep{lightman2023let}. Consequently, enhancing LLM capabilities at inference time has become a critical research frontier \citep{scaling}. A prominent direction involves leveraging search algorithms, particularly reward-guided tree search, to improve complex reasoning \citep{ke2025survey}. These methods typically expand the search space by generating multiple reasoning steps (e.g., via chain-of-thought prompting \citep{wei2022chain}) and 
employ various techniques to navigate this space, from foundational algorithms like depth-first search \citep{yao2023tree} to more advanced heuristics like beam search \citep{kang2024mindstar} and Monte Carlo Tree Search (MCTS) \citep{chenalphamath,zhang2024rest}, to identify high-quality solutions.

However, current search strategies exhibit significant inefficiencies \citep{damani2024learning}. A common practice involves sampling multiple candidate reasoning steps 
and exploring them, often allocating computational resources uniformly across these candidates \citep{yao2023tree,long2023large,besta2024graph}. This approach overlooks the potential semantic equivalence among textually distinct candidates. Treating semantically identical reasoning steps as unique branches leads to redundant exploration of the search space, incurring substantial computational overhead through unnecessary token generation and processing. Moreover, for search algorithms that incorporate preference learning based on intermediate steps (e.g., \citealp{xie2024monte,jiang2024technical}), learning preferences from pairs of equivalent steps may provide noisy or conflicting signals, hindering the learning of effective reasoning policies. This challenge is particularly acute in domains like mathematical reasoning, where numerous textual formulations can represent the same underlying logical operation or state.
\begin{figure}[t]
  \includegraphics[width=\linewidth]{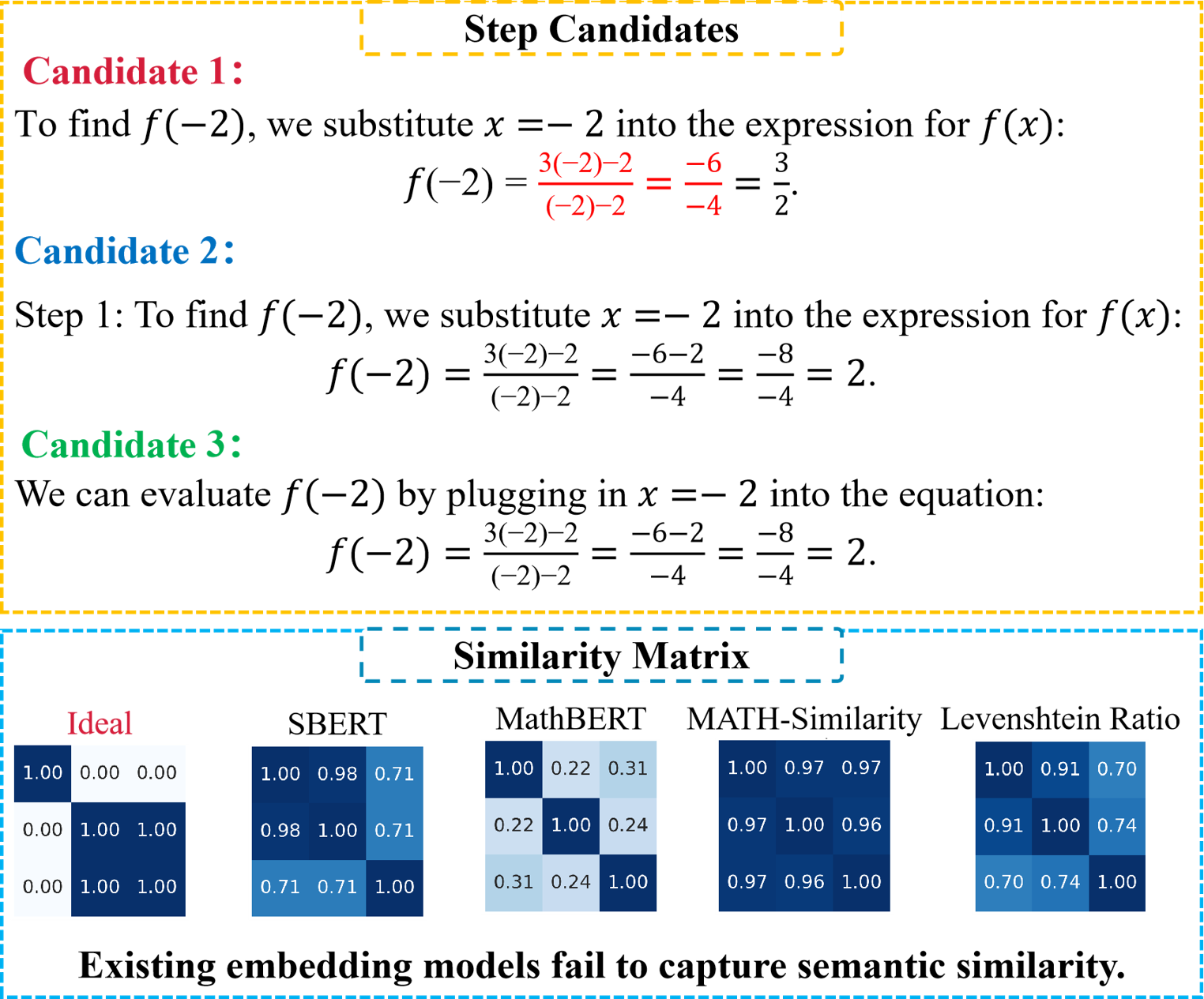}
    \caption{Illustration of the mathematical statement equivalence challenge during reasoning search. Given multiple candidate steps generated by an LLM, standard methods like embedding similarity or Levenshtein Ratio may incorrectly assess candidate 1 and candidate 2 as highly similar due to surface features, while failing to recognize the true semantic equivalence between candidate 2 and candidate 3, which represent the identical logical operation. }
    \label{fig:intro}
\end{figure}
Addressing this redundancy via standard Semantic Textual Similarity (STS) techniques \citep{majumder2016semantic} proves inadequate as illustrated in Figure \ref{fig:intro}. Existing embedding models, such as SBERT \citep{reimers2019sentence}, predominantly trained on general text, often fail to capture the nuanced structural and logical equivalence specific to mathematical statements. Even domain-specific models like MathBERT \citep{peng2021mathbert}, which enhance mathematical text representation, along with other embedding models MATH-Similarity \citep{steinfeldt2024evaluation}, lack optimization for identifying functional equivalence between mathematical sentences. This limitation is further exacerbated by the lack of specialized benchmark datasets designed for mathematical statement equivalence. Although large-scale generative models can achieve satisfactory performance in few-shot scenarios for such judgment tasks, their substantially higher computational complexity results in significantly slower inference speeds compared to embedding models \citep{brown2020language}. The consequent latency renders them impractical for high-throughput applications requiring real-time processing.

To overcome these limitations, we introduce \textit{EquivPruner}, a simple yet effective approach that centers on identifying and pruning semantically equivalent actions during LLM reasoning search. 
We create MathEquiv, the first dataset specifically designed for mathematical statement equivalence. 
Leveraging this dataset, we trained a lightweight yet effective equivalence detection model. This model serves as a dynamic pruner integrated into the LLM's search process. When the LLM generates multiple candidate reasoning steps at a given expansion point, the pruner identifies sets of semantically equivalent candidates among these siblings. For each set of equivalent steps, it retains only a single representative node for further exploration, effectively pruning the redundant branches and significantly reducing the search space.

While the proposed pruning framework is potentially generalizable, this paper focuses on its validation within mathematical reasoning due to the significant research community attention \citep{ke2025survey} and the availability of well-developed open-source process reward models \citep{shao2024deepseekmath}.
We conduct extensive experiments across various models, including Mistral-7B-SFT \citep{shao2024deepseekmath} and the Qwen2.5-Math-Instruct series \citep{yang2024qwen2}, using two widely recognized math reasoning benchmarks: GSM8K \citep{cobbe2021training} and MATH \citep{hendrycks2021math}.
Our proposed EquivPruner demonstrates compelling improvements across these settings. For instance, when applied to Qwen2.5-Math-7B-Instruct on GSM8K—where the model already achieves a very high baseline accuracy of 96.44\%—EquivPruner not only cuts token consumption by a substantial 48.1\% but also further boosts accuracy to 96.59\%. This demonstrates EquivPruner's ability to significantly enhance searching efficiency.

Our main contributions are:
\begin{itemize}
\item To the best of our knowledge, this work is the \textit{first} to identify and address the problem of action equivalence in LLM-based reasoning search.
\item We introduce \textit{EquivPruner}, a simple yet effective approach that centers on identifying and pruning semantically equivalent actions during LLM reasoning search.
\item We release MathEquiv, the first benchmark dataset specifically designed for mathematical statement equivalence. It serves as a versatile resource applicable to a variety of mathematical tasks and scenarios.
\item Extensive experiments demonstrate the effectiveness of \textit{EquivPruner}.
When applied to Qwen2.5-Math-7B-Instruct on GSM8K, \textit{EquivPruner} not only cuts token consumption by a substantial 48.1\% but also further boosts accuracy in a very high baseline.
\end{itemize}

\section{Related Work}
\paragraph{LLM Reasoning via Search Strategies}
Efforts to improve LLM problem-solving capabilities have moved beyond simple prompting. Chain-of-Thought prompting \citep{wei2022chain} demonstrated the value of intermediate reasoning steps. Building on this, structured search methods like Tree-of-Thoughts \citep{yao2023tree} and Graph-of-Thoughts \citep{besta2024graph} explore multiple reasoning paths, enhancing performance on complex tasks requiring exploration and backtracking.
Further advancing this direction, a particularly powerful paradigm integrates LLMs with sophisticated search algorithms. Among these, the synergy between LLMs and Monte Carlo Tree Search (MCTS) \citep{chenalphamath,zhang2024rest} has garnered significant attention for tackling complex reasoning problems.
MCTS, renowned for its ability to balance exploration and exploitation in vast search spaces, becomes exceptionally potent when guided by an LLM's generative capabilities to propose candidate steps and a reward model to estimate state values \citep{yao2023tree,long2023large,besta2024graph}. This LLM-MCTS approach, alongside other advanced search integrations like LLM-guided beam search \citep{chenalphamath} and depth-first-search \citep{yao2023tree}, has consistently achieved state-of-the-art results in demanding areas such as science tasks \citep{yang2024qwen2}, coding \citep{dainesegenerating,zhangplanning}, and mathematical reasoning \citep{zhang2024llama, luo2024improve}.
However, despite the remarkable success of these advanced search strategies, a significant challenge emerges, especially prevalent in mathematical reasoning when employing methods like LLM-MCTS: the substantial token cost \citep{chenalphamath}. While LLM-MCTS explores many branches effectively, it often wastes resources evaluating syntactically distinct but semantically equivalent states. This redundancy unnecessarily expands the search space, consuming tokens without yielding novel solutions, thus limiting efficiency and scalability.
\paragraph{Mathematical Equivalence Detection}
Effective detection of mathematical statement equivalence is crucial for efficient LLM-Based search tree pruning, yet current methodologies exhibit significant shortcomings. For instance, rudimentary sequence comparison metrics like Levenshtein similarity \citep{yujian2007normalized} are fundamentally ill-suited, as they fail to capture the deep semantic and hierarchical structures inherent in mathematical language, leading to unreliable equivalence assessments.
% Levenshtein similarity, a common sequence comparison metric, fails to account for the semantic and hierarchical nature of mathematical language, leading to inaccurate equivalence detection. 
Standard Semantic Textual Similarity models, such as SBERT \citep{reimers2019sentence}, trained predominantly on general language corpora, are designed to capture semantic relatedness rather than strict mathematical equivalence. Even domain-specific models like MathBERT \citep{peng2021mathbert}, which enhance mathematical text representation, along with other embedding models MATH-Similarity \citep{steinfeldt2024evaluation}, lack optimization for identifying functional equivalence between mathematical sentences. 
Their capacity to accurately recognize semantically equivalent mathematical sentences is thereby constrained, as illustrated by the examples in Figure \ref{fig:intro}.
While LLMs like GPT-4o \citep{hurst2024gpt} has the ability to recognize mathematical equivalence, their complex architectures introduce significant latency.
This high time overhead renders them impractical for real-time pruning scenarios.
Consequently, there is an urgent need to enable efficient pruning in LLM-based search.
\begin{figure*}[t]
  \includegraphics[width=\linewidth]{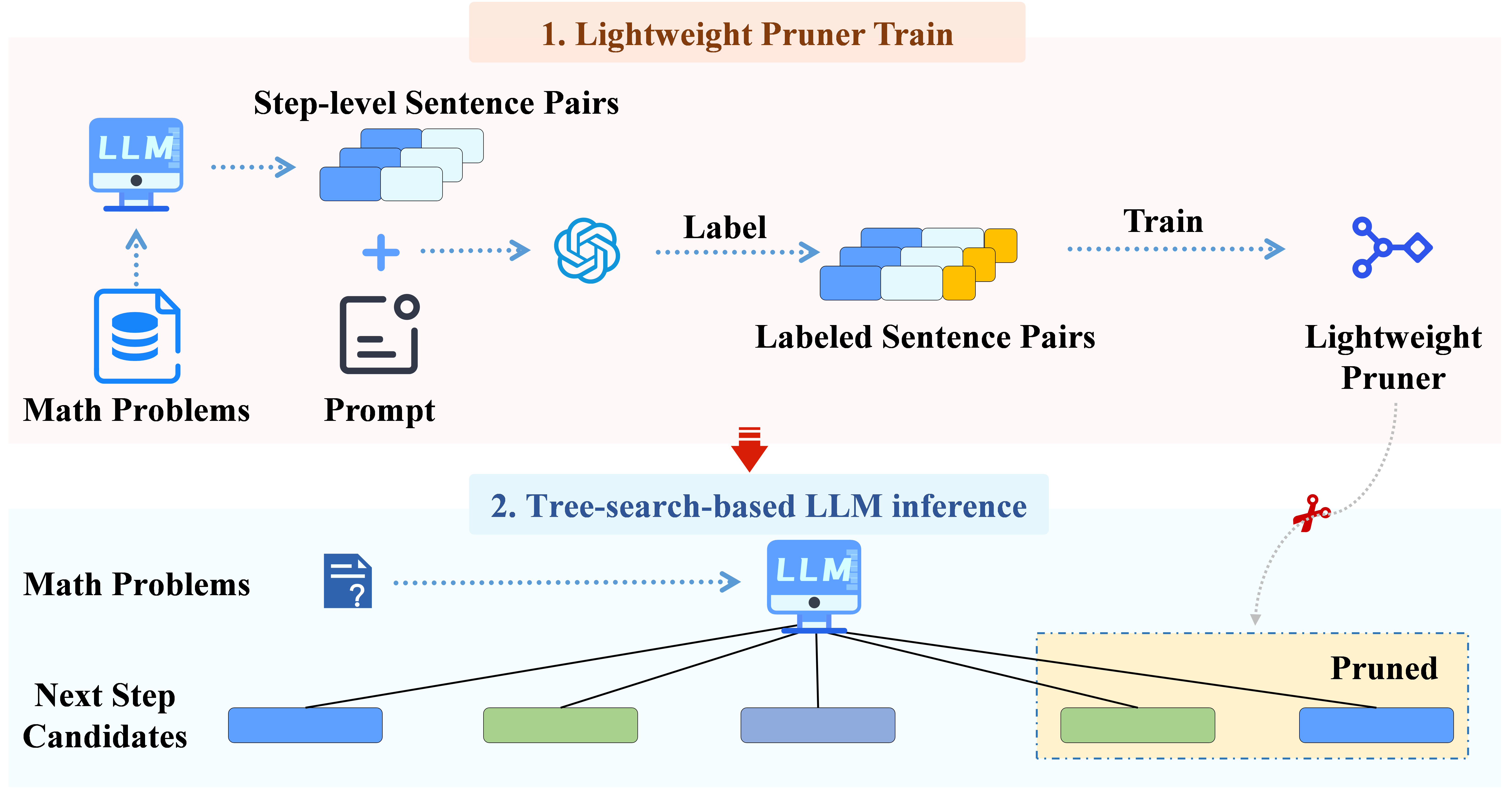}
  \caption{The EquivPruner framework. Top: Training the lightweight equivalence pruner from labeled step-level sentence pairs. Bottom: Applying the trained lightweight pruner during tree-search-based LLM inference to remove semantically equivalent candidates generated by the LLM.}
  \label{fig:experiments}
\end{figure*}
\section{Methodology}
\label{headings}

\subsection{Define Semantic Equivalence in Mathematics}
Simply equating statements based on identical outcomes can be superficial and misleading, as it may overlook critical differences in conceptual articulation, structural formulation, symbolic interpretation, and methodological pathways. To address this, we propose a definition of semantic equivalence specifically attuned to these multifaceted aspects. Accordingly, in our framework, two mathematical statements are considered semantically equivalent if and only if they rigorously satisfy the following criteria:
\begin{itemize}
\item \textbf{Conceptual Consistency:} The statements must articulate identical mathematical concepts, definitions, or propositions without ambiguity.
\item \textbf{Structural Equivalence:} Their logical formulations, encompassing assumptions, derivations, and conclusions, must be fully aligned.
\item \textbf{Notational Precision:} All variables, symbols, and mathematical expressions must be employed consistently, maintaining identical meanings across the statements.
\item \textbf{Methodological Congruence:} Semantic equivalence necessitates an alignment in the underlying methodology and reasoning. Statements yielding the same final result via disparate approaches are not considered fully equivalent.
\end{itemize}

Our approach to semantic equivalence thus mandates a comprehensive assessment. It scrutinizes the congruence of conceptual foundations, logical structures, notational usage, and methodological approaches. Two mathematical statements are judged completely equivalent only when they demonstrate indivisible identity across all these critical facets.

\subsection{The MathEquiv Dataset}
Recognizing the absence of dedicated datasets for mathematical statement equivalence, we constructed and released MathEquiv to bridge this gap.
The MathEquiv dataset was curated by initially employing a Step-level Beam Search algorithm \citep{chenalphamath} to gather action candidates. These candidates were subsequently formulated into step-level sentence pairs.

For the task of equivalence scoring, we implemented a five-tiered classification system. This granular approach was adopted to enhance the stability of the GPT model's outputs, as preliminary experiments with binary classification (equivalent/non-equivalent) revealed inconsistencies in judgments. The five-tiered system yielded significantly more consistent and reliable assessments:
\begin{itemize}
\item \textbf{Level 4 (Exactly Equivalent):} The statements are mathematically interchangeable in all respects, exhibiting identical meaning and form.
\item \textbf{Level 3 (Likely Equivalent):} Minor syntactic differences may be present, but the core mathematical content and logic align.
\item \textbf{Level 2 (Indeterminable):} Insufficient information is available to make a definitive judgment regarding equivalence.
\item \textbf{Level 1 (Unlikely Equivalent):} While some partial agreement may exist, critical discrepancies in logic, definition, or mathematical structure are observed.
\item \textbf{Level 0 (Not Equivalent):} The statements are fundamentally distinct in their mathematical meaning, derivation, or resultant outcomes.
\end{itemize}

The MathEquiv dataset was labeled via an iterative refinement process. 
% Initially, GPT-4o labeled a data subset, followed by human expert review. For discrepancies, the human-adjudicated label and its rationale were incorporated into GPT-4o's prompt as few-shot examples. This cycle was repeated until model outputs for a randomly sampled subset consistently aligned with human consensus. Subsequently, the collection of few-shot examples was pruned to a minimal, representative set sufficient to maintain this model-human alignment. 
We began selecting few-shot examples by randomly sampling 20 instances and obtaining independent labels from both human and GPT-4o. When discrepancies occurred, the human-adjudicated label and its rationale were incorporated into GPT-4o's prompt as few-shot examples, and the process was repeated. 
After three such iterations, we observed stable agreement between human and GPT annotations. Upon analyzing those few-shot samples, we identified two dominant failure modes where the model was overly permissive in assigning equivalent label. The first is that GPT focuses solely on the final mathematical conclusion, without accounting for the conditions. And the second is that GPT labeles equivalence when they reached the same answer, even if the intermediate mathematical procedures were different.
We incorporated representative non-equivalent examples from these two categories into the final few-shot prompt. After this refinement, we evaluated a new random set of 100 samples and observed complete agreement between human and GPT annotations.
This process yielded the MathEquiv dataset, characterized by high-quality labels and an accurate assessment of semantic equivalence. The final prompt is detailed in Figure~\ref{prompt}. We show examples of the contents of MathEquiv in Figure \ref{4example} and Figure \ref{1example}. The MathEquiv dataset is available at \url{https://huggingface.co/datasets/Jiawei1222/MathEquiv}.

\subsection{Lightweight Pruner for Tree Search}
To facilitate dynamic, real-time pruning within our tree search algorithm, we developed and trained a dedicated Lightweight Pruner. The data collection process for training this pruner and its integration into the broader Tree-search-based LLM inference pipeline are illustrated in Figure 2.

\subsubsection{Data Complexity in Pruner Training}
The MathEquiv dataset, suitable for assessing overall statement equivalence, presents specific challenges for training the Lightweight Pruner. The dataset's step-level sentence pairs often consist of multiple sentences. A key difficulty is that step pairs labeled as non-equivalent at a macro-level may nevertheless contain sub-pairs of sentences that are semantically equivalent. This characteristic, common in data derived from intermediate mathematical problem-solving steps, can introduce ambiguity and hinder the pruner's ability to learn fine-grained distinctions if not appropriately addressed. The true equivalence status of these sub-sentence pairs can be viewed as a latent aspect of the data.

\subsubsection{Pruner Training via Expectation-Maximization (EM)}
To effectively train the Lightweight Pruner amidst this data complexity, we employ the Expectation-Maximization (EM) algorithm, which is effective for handling the unobserved equivalence status of sub-sentence pairs within larger, complex training instances. 
The algorithm iteratively maximizes the expected complete-data log-likelihood by alternating between two steps:

\textbf{1. E-step (Expectation Step):}
Given the model parameters $\theta^{(t)}$ at iteration $t$, the pruner predicts the equivalence probability of each sub-sentence pair in multi-sentence samples. Sub-sentence pairs with probabilities exceeding a threshold are treated as high-confidence equivalents and removed from samples to refine the dataset for the next step.

\textbf{2. M-step (Maximization Step):}
The model parameters are updated to $\theta^{(t+1)}$ by maximizing the likelihood of the observed data, conditioned on the expectations derived in the E-step.

By training on samples that have been simplified or where latent equivalences have been accounted for, the model can better focus on learning more subtle or challenging distinctions necessary for effective pruning.

\subsection{Pruner's Impact on Different Search Algorithms}
The pruner enhances search performance across various algorithms by eliminating state redundancy, with its contribution tailored to the architecture of each search strategy. For heuristic-driven algorithms like Monte Carlo Tree Search (MCTS), it optimizes the allocation of the computational budget (i.e., simulation rollouts) by collapsing equivalent states. This concentrates exploration on unique paths, which improves the accuracy of the guiding policy statistics and accelerates convergence. In Step-level Beam Search (SBS), the pruner directly improves search quality by diversifying the beam. By removing equivalent nodes \textit{prior} to the top-$k$ selection, it ensures that the beam's limited slots are occupied by unique candidates, mitigating the risk of optimal paths being prematurely pruned. For systematic algorithms such as Depth-First Search (DFS), it yields a direct efficiency gain by preventing the repeated exploration of identical subtrees.
% The pruner's primary benefit is its synergistic interaction with the heuristic nature of MCTS. By merging equivalent child nodes, it prevents wasting the finite computational budget (i.e., simulation rollouts) on redundant states. This allows MCTS to concentrate its search resources on unique pathways, yielding more accurate statistics to guide its UCT policy. This focused exploration enhances the intelligence of the search, leading to both greater efficiency and accelerated convergence to optimal solutions.

% In contrast, for simpler systematic algorithms, the pruner primarily enhances efficiency and search quality. For Depth-First-Search (DFS), it provides a crucial efficiency gain by preventing the redundant traversal of identical subtrees. For Step-level Beam Search (SBS), its role is more critical: it ensures the fixed-width beam (k) is composed of distinct states. This maximizes search diversity and prevents the optimal path from being prematurely discarded due to redundant candidates occupying valuable slots.
% For Step-level Beam Search (SBS), the pruner is critical for search quality. By eliminating equivalent nodes before the top-k selection at each step, it ensures the fixed-width beam is composed of distinct candidates. This maximizes search diversity and prevents the optimal path from being prematurely discarded due to redundant states occupying valuable slots.
% For systematic algorithms like Depth-First Search (DFS), it provides a crucial efficiency gain by preventing the redundant traversal of identical subtrees.
\begin{table*}[ht]
\caption{Performance comparison of Vanilla MCTS and MCTS + EquivPruner across three language models on the MATH and GSM8K datasets. EquivPruner significantly reduces token consumption (Tokens, Ratio) while generally maintaining or improving accuracy (Acc, \%). Best results within each model-dataset block are in \textbf{bold}.}
\label{maintable}
\centering
\begin{tabular}{ccccccc}
\hline
\multirow{2}{*}{Methods}                                            & \multicolumn{3}{c}{MATH}                           & \multicolumn{3}{c}{GSM8K}                          \\ \cline{2-7} 
                                                                  & Acc            & Tokens         & Ratio            & Acc            & Tokens         & Ratio            \\ \hline
\multicolumn{1}{l}{\textit{\textbf{Qwen2.5-Math-7B-Instruct:}}}   &                &                &                  &                &                &                  \\
Vanilla MCTS                                                      & 83.40          & 106773         & 100.00\%         & 96.44          & 34826          & 100.00\%         \\
\multicolumn{1}{r}{+ EquivPruner}                                 & \textbf{84.00} & \textbf{74194} & \textbf{69.49\%} & \textbf{96.59} & \textbf{18071} & \textbf{51.89\%} \\ \hline
\multicolumn{1}{l}{\textit{\textbf{Mistral-7b-sft:}}}             &                &                &                  &                &                &                  \\
Vanilla MCTS                                                      & 36.60          & 49251          & 100.00\%         & 83.78          & 20217          & 100.00\%         \\
\multicolumn{1}{r}{+ EquivPruner}                                 & \textbf{37.40} & \textbf{38265} & \textbf{77.69\%} & \textbf{85.06} & \textbf{12537} & \textbf{62.01\%} \\ \hline
\multicolumn{1}{l}{\textit{\textbf{Qwen2.5-Math-1.5B-Instruct:}}} &                &                &                  &                &                &                  \\
Vanilla MCTS                                                      & 75.60          & 91811          & 100.00\%         & \textbf{91.05}          & 39337          & 100.00\%         \\
\multicolumn{1}{r}{+ EquivPruner}                                 & \textbf{75.60} & \textbf{71878} & \textbf{78.29\%} & 90.75 & \textbf{23752} & \textbf{60.38\%} \\ \hline
\end{tabular}
\end{table*}

\section{Experiments}
\label{sec:Experiments}

In this section, we present a series of comprehensive experiments designed to validate the efficacy of  EquivPruner.

\subsection{MathEquiv Dataset Generation}
\label{subsec:MathEquiv_Dataset_Generation}

We constructed the MathEquiv dataset for mathematical statement equivalence. The foundation of this dataset consists of 7,500 mathematical problems sourced from the MATH training set \citep{hendrycks2021math}. To prevent data leakage between training, validation, and test phases of EquivPruner, these 7,500 problems were first split into training, validation, and test sets using an 8:1:1 ratio, respectively. For each problem in these distinct sets, we generated candidate reasoning step pairs using the Qwen2.5-Math-7B-Instruct model \citep{yang2024qwen2} via Step-level Beam Search. These pairs were subsequently filtered based on Levenshtein distance, and a balanced sample from each set was then annotated for equivalence by GPT-4o. This process resulted in distinct training, validation, and test sets of annotated step pairs for EquivPruner. The specific parameters for step pair generation, filtering criteria, and the final dataset sizes are detailed in Appendix \ref{app:Dataset_Generation_Details}.

\subsection{Experimental Setup}
\label{subsec:Experimental_Setup}

\subsubsection{Models and Datasets}
\label{subsubsec:Models_and_Datasets}

For inference, we utilized several LLMs: Qwen2.5-Math-7B-Instruct \citep{yang2024qwen2}, Mistral-7B-SFT \citep{shao2024deepseekmath}, Qwen2.5-Math-1.5B-Instruct \citep{yang2024qwen2}, and Qwen2.5-14B-Instruct~\citep{yang2024qwen2}. 
Given that existing open-source PRMs are predominantly tailored for mathematical reasoning, our current investigation is confined to mathematical tasks. Nevertheless, the EquivPruner framework is designed for generalizability and can be readily extended to other domains like code generation and commonsense reasoning. 
The Process Reward Model (PRM) employed for guiding the Monte Carlo Tree Search (MCTS) was Math-Shepherd-Mistral-7B-PRM \citep{shao2024deepseekmath}. As EquivPruner was trained on data generated by Qwen2.5-Math-7B-Instruct, the Mistral-7B-SFT and Qwen2.5-Math-1.5B-Instruct models serve as out-of-distribution (OOD) models in our experiments.

Our evaluation was mainly conducted on three standard benchmark datasets:
\begin{itemize}
\item \textbf{MATH} \citep{hendrycks2021math}: Featuring challenging competition-level mathematics problems. Due to computational demands, our evaluation on the MATH dataset was performed on the MATH-500 subset, identical to the test partition used in \citet{lightman2023let}.
\item \textbf{GSM8K} \citep{cobbe2021training}: Consisting of grade school mathematics word problems. Its test set has 1319 problems. Since EquivPruner was trained on data derived from MATH dataset problems, GSM8K is considered an OOD dataset.
\item \textbf{AIME 2024} \citep{codeforcesamerican}: The AIME 2024 dataset consists of 30 challenging mathematics problems spanning various domains, including algebra, geometry, and number theory. AIME 2024 is also considered an OOD dataset.
\end{itemize}
We also conducted additional cross-domain experiments on HumanEval~\cite{chen2021evaluating} and CommonsenseQA~\cite{talmor2019commonsenseqa} datasets.
\subsubsection{Implementation Details}
\label{subsubsec:Implementation_Details}

The EquivPruner model itself is a fine-tuned Longformer-base \citep{beltagy2020longformer}, chosen for its efficiency suitable for real-time pruning.
We report evaluation results of pruner in Appendix \ref{eval}. During the MCTS inference phase, the determination of equivalence between two reasoning step nodes involves a two-stage process. First, the Levenshtein ratio between the steps is calculated. If the ratio is less than or equal to 0.75, the nodes are immediately considered non-equivalent, acting as a fast filter. Only if the Levenshtein ratio is greater than 0.75 is the EquivPruner model invoked to make the final equivalence prediction. This hierarchical check balances speed and accuracy in the pruning process.
The maximum number of newly generated tokens by the LLMs (max\_new\_tokens) was set to 1024, and the generation temperature was 0.7.  All experiments were conducted on NVIDIA GeForce RTX 3090 GPUs. 
Further details are available in Appendix \ref{app:Implementation_Environment}.

\subsection{Evaluation Metrics}
\label{subsec:Evaluation_Metrics}

We adopted a vanilla MCTS \citep{chenalphamath} as the baseline for comparison. The evaluation of EquivPruner focuses on two primary aspects:
\begin{itemize}
\item \textbf{Effectiveness}: Measured using solution accuracy (Acc), the percentage of problems solved correctly.
\item \textbf{Efficiency}: Assessed through the average number of tokens generated (Tokens) and a token ratio (Ratio), defined as the ratio of tokens generated by the EquivPruner-enhanced search to those generated by the baseline MCTS.
\end{itemize}

\subsection{Main Results}
\label{subsec:Main_Results}
Table \ref{maintable} presents our main experimental findings, comparing vanilla MCTS against MCTS augmented with EquivPruner on MATH dataset and GSM8K dataset. We also conducted experiments on AIME 2024 dataset in the Appendix \ref{app:aime}. The results consistently demonstrate that EquivPruner substantially enhances computational efficiency across different language models and datasets, primarily by reducing token generation while largely preserving or even improving solution accuracy.
Due to page limitations, the failure mode analysis is provided in Appendix~\ref{fma}.
\paragraph{Efficiency Gains}
EquivPruner achieves significant reductions in token counts across all configurations. For instance, with Qwen2.5-Math-7B-Instruct on GSM8K, tokens were reduced by approximately 48.11\% (Ratio: 51.89\%), and on MATH, by 30.51\% (Ratio: 69.49\%). Similar substantial token savings were observed for Mistral-7B-SFT (e.g., ~37.99\% reduction on GSM8K) and Qwen2.5-Math-1.5B-Instruct (e.g., ~39.62\% reduction on GSM8K). These figures highlight EquivPruner's effectiveness in pruning the search space.

\paragraph{Accuracy Impact and Resource Optimization}
Crucially, these efficiency improvements are generally accompanied by maintained or enhanced accuracy. Qwen2.5-Math-7B-Instruct saw accuracy gains of +0.60\% on MATH and +0.15\% on GSM8K. With Mistral-7B-SFT, an OOD model relative to EquivPruner's training data source, accuracy improved by +0.80\% on MATH and +1.28\% on GSM8K (also an OOD dataset for EquivPruner). This suggests that by eliminating redundant explorations, EquivPruner enables MCTS to allocate its search resources more effectively. For Qwen2.5-Math-1.5B-Instruct (another OOD model), accuracy was maintained on MATH and saw a minor dip of -0.30\% on GSM8K, which is a reasonable trade-off given the nearly 40\% token reduction.

\paragraph{Generalization}
The positive outcomes on OOD models (Mistral-7B-SFT, Qwen2.5-Math-1.5B-Instruct) and the OOD dataset (GSM8K) underscore EquivPruner's generalization capabilities. It effectively identifies and removes equivalent reasoning steps, allowing MCTS to conduct a more focused and efficient search across varied settings.

\begin{table}[t]
\centering
\caption{Performance comparison of SBERT, MATH-Similarity, MathBERT, Levenshtein ratio (threshold = 0.95), and EquivPruner for pruning across two language models on the MATH datasets. EquivPruner significantly reduces token consumption (Tokens) while improving accuracy (Acc, \%).}
\label{tab:baseline}
\begin{tabular}{rcc}
\hline
\multicolumn{1}{c}{Methods}                                     & Acc                  & Tokens               \\ \hline
\multicolumn{1}{l}{\textit{\textbf{Qwen2.5-Math-7B-Instruct:}}} & \multicolumn{1}{l}{} & \multicolumn{1}{l}{} \\
\multicolumn{1}{c}{Vanilla MCTS}                                & 83.40                & 106773               \\
+SBERT Pruner                                                   & 82.60                & 70142                \\
+MATH-Similarity Pruner                                         & 82.00                & 76693                \\
+MathBERT Pruner                                         & 83.80                & 76936                \\
+Levenshtein Ratio Pruner                                       & 83.00                & 105118               \\
+EquivPruner (Ours)                                             & 84.00                & 74194                \\ \hline
\multicolumn{1}{l}{\textit{\textbf{Mistral-7b-sft:}}}           & \multicolumn{1}{l}{} & \multicolumn{1}{l}{} \\
\multicolumn{1}{c}{Vanilla MCTS}                                & 36.60                & 49251                \\
+SBERT Pruner                                                   & 33.00                & 30876                \\
+MATH-Similarity Pruner                                         & 34.40                & 33212                \\
+Levenshtein Ratio Pruner                                       & 35.50                & 46608                \\
+EquivPruner(Ours)                                              & 37.40                & 38265                \\ \hline
\end{tabular}
\end{table}
\subsection{Baseline Comparison with Alternative Pruning Methods}
To further evaluatethe effectiveness of our proposed EquivPruner, we compare it against alternative pruning methods based on semantic similarity (SBERT, MATH-Similarity, MathBERT) and string similarity (Levenshtein Ratio, threshold = 0.95). We also explored the Formula Extraction-based method in the Appendix \ref{app:Formula_Extraction} considering the challenges in consistently parsing mathematical expressions.

Table \ref{tab:baseline} shows that these baselines are suboptimal. Embedding-based pruners reduce tokens but at the cost of accuracy; for instance, they caused an accuracy drop of up to 3.6\% with the Mistral-7B-sft model. This indicates semantic similarity is a poor proxy for mathematical equivalence, leading to the erroneous pruning of valid reasoning paths. Similarly, the Levenshtein Ratio pruner, while maintaining accuracy, provided a negligible token reduction of only 1.5\% with the Qwen model.

In contrast, our EquivPruner improves both efficiency and performance. It achieved a 30.4\% token reduction with Qwen2.5-Math-7B-Instruct while also increasing accuracy on both models (e.g., from 36.6\% to 37.4\% on Mistral-7B-sft). This confirms that checking for functional equivalence is superior to relying on semantic or textual similarity for pruning.

\begin{table}[t]
\centering
\caption{Performance of EquivPruner with Step-level Beam Search (SBS) using the Qwen2.5-Math-7B-Instruct model on MATH, GSM8K and AIME 2024. EquivPruner enhances accuracy (Acc, \%) by promoting diversity among selected nodes, with token counts (Tokens, Ratio) remaining largely stable.}
\label{tab:fur}
\begin{tabular}{cccc}
\hline
Methods                                      & Acc   & Tokens & Ratio    \\ \hline
\multicolumn{1}{l}{\textit{\textbf{MATH:}}}  &       &        &          \\
SBS                       & 82.00 & 21341  & 100.00\% \\
\multicolumn{1}{r}{+ EquivPruner}            & 82.20 & 20952  & 98.18\%  \\ \hline
\multicolumn{1}{l}{\textit{\textbf{GSM8K:}}} &       &        &          \\
SBS                       & 96.06 & 8004  & 100.00\% \\
\multicolumn{1}{r}{+ EquivPruner}            & 96.13 & 7927  & 99.04\%    \\ \hline
\multicolumn{1}{l}{\textit{\textbf{AIME 2024:}}} &       &        &          \\
SBS                       & 10.00 & 46672  & 100.00\% \\
\multicolumn{1}{r}{+ EquivPruner}            & 13.33 & 43391  & 92.97\%    \\ \hline
\end{tabular}
\end{table}

\subsection{Effectiveness in Step-level Beam Search}
\label{subsec:EquivPruner_SBS}
To empirically validate the pruner's benefit for SBS, we integrated it into the search process using the Qwen2.5-Math-7B-Instruct model.

The results in Table \ref{tab:fur} confirm its positive impact on search quality. EquivPruner increased accuracy on MATH from 82.00\% to 82.20\% (+0.20\%), on GSM8K from 96.06\% to 96.13\% (+0.07\%), and on AIME 2024 from 10.00\% to 13.33\% (+3.33\%). As hypothesized, these gains were achieved without significant changes in token generation, with ratios of 98.18\% (MATH), 99.04\% (GSM8K), and 92.97\% (AIME 2024).
These findings show that by ensuring the top-$k$ candidates are distinct, the pruner fosters a more diverse and effective exploration of the solution space, validating its role as a versatile component for improving LLM reasoning search strategies.
\subsection{Effectiveness in Lookahead Search}
We integrated EquivPruner with Lookahead Search~\cite{snell2025scaling} to evaluate its compatibility. Using Qwen2.5-Math-7B-Instruct on the MATH dataset, EquivPruner improves accuracy from 80.8\% to 82.6\% while maintaining comparable token efficiency (Table~\ref{tab:lookahead}). This indicates that the pruner promotes diversity among top-k candidates, enabling more effective exploration of the solution space. Due to page limitations, the section "Effectiveness in Depth-First-Search" is provided in Appendix \ref{dfsapp}.
\begin{table}[t]
\centering
\caption{
Performance of EquivPruner with Lookahead Search using the Qwen2.5-Math-7B-Instruct model on MATH.}
\label{tab:lookahead}
\begin{tabular}{cccc}
\hline
Methods                                      & Acc                  & Tokens                        & Ratio                \\ \hline
\multicolumn{1}{l}{\textit{\textbf{MATH:}}}  & \multicolumn{1}{l}{} & \multicolumn{1}{l}{\textbf{}} & \multicolumn{1}{l}{} \\
Lookahead Search                                          & 80.80                & 20131                & 100.00\%             \\
\multicolumn{1}{r}{+EquivPruner}             & 82.60                & 19978                & 99.24\%              \\ \hline
\end{tabular}
\end{table}
\subsection{Effectiveness in Cross-Domain Tasks}
To evaluate the generalizability of EquivPruner, we conducted additional experiments on code generation (HumanEval) and commonsense reasoning (CommonsenseQA, 100 random samples) using Qwen2.5-Math-7B-Instruct. As shown in Table~\ref{tab:cross}, EquivPruner consistently improves accuracy while reducing token consumption across both domains. These results confirm that pruning equivalent search states is broadly effective beyond mathematical reasoning tasks.
\begin{table}[t]
\centering
\caption{Performance of EquivPruner with Vanilla MCTS and MCTS + EquivPruner using the Qwen2.5-Math-7B-Instruct model on Code Generation Task (HumanEval Dataset) and Commonsense Reasoning Task (CommonsenseQA Dataset).}
\label{tab:cross}
\begin{tabular}{cccc}
\hline
Methods                                      & Acc                  & Tokens                        & Ratio                \\ \hline
\multicolumn{1}{l}{\textit{\textbf{HumanEval:}}}  & \multicolumn{1}{l}{} & \multicolumn{1}{l}{\textbf{}} & \multicolumn{1}{l}{} \\
Vanilla MCTS                                          & 19.51                & 767                & 100.00\%             \\
\multicolumn{1}{r}{+EquivPruner}             & 29.27                & 737                & 96.09\%              \\ \hline
\multicolumn{1}{l}{\textit{\textbf{CommonsenseQA:}}} & \multicolumn{1}{l}{} & \multicolumn{1}{l}{}          & \multicolumn{1}{l}{} \\
Vanilla MCTS                                          & 56.0                & 39827                         & 100.00\%             \\
\multicolumn{1}{r}{+EquivPruner}             & 57.0                & 35129                          & 88.20\%              \\ \hline
\end{tabular}
\end{table}
\subsection{Scalability to Larger Models}
We further evaluated EquivPruner on the larger Qwen2.5-14B-Instruct model using a random subset of 100 questions from GSM8K. As shown in Table~\ref{tab:gsm8k-14b}, the results align with our findings on smaller models: EquivPruner reduces token consumption to 61.71\% of the baseline while improving accuracy from 96.0\% to 97.0\%. This confirms that our approach scales effectively to larger models.
\begin{table}[t]
\centering
\caption{Performance of EquivPruner with Vanilla MCTS and MCTS + EquivPruner using the Qwen2.5-14B-Instruct model on GSM8K.}
\label{tab:gsm8k-14b}
\begin{tabular}{cccc}
\hline
Methods                                      & Acc                  & Tokens                        & Ratio                \\ \hline
\multicolumn{1}{l}{\textit{\textbf{GSM8K:}}}  & \multicolumn{1}{l}{} & \multicolumn{1}{l}{\textbf{}} & \multicolumn{1}{l}{} \\
Vanilla MCTS                                          & 96.0                & 44404                & 100.00\%             \\
\multicolumn{1}{r}{+EquivPruner}             & 97.0                & 27402                & 61.71\%              \\ \hline
\end{tabular}
\end{table}
\subsection{Ablation Study}
\label{ablation}
\begin{figure}[t]
  \includegraphics[width=\linewidth]{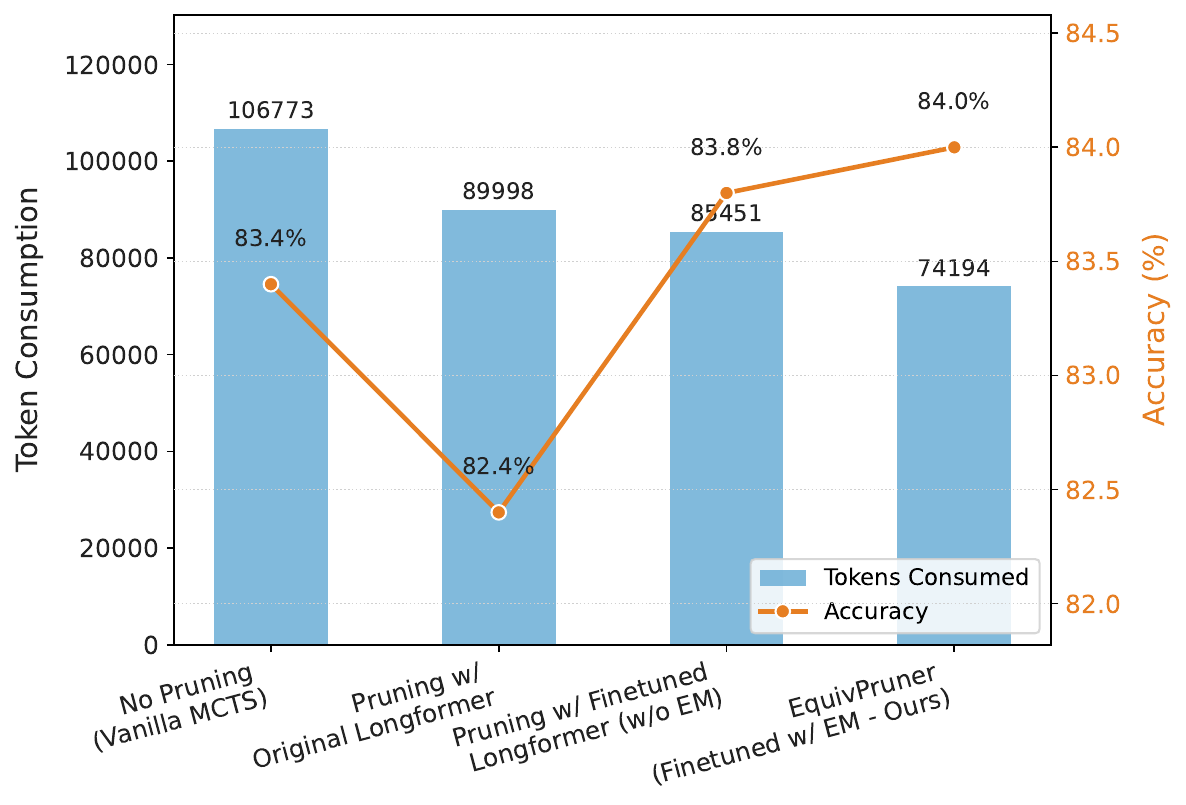}
    \caption{Ablation study of EquivPruner components. The plot illustrates the impact of different pruning strategies within a MCTS framework on Token Consumption (bars, left y-axis) and Accuracy (line, right y-axis). }
    \label{fig:ablation}
\end{figure}

To investigate the individual contributions of the key components of our EquivPruner—specifically, the fine-tuning process and the use of the EM algorithm—we conducted an ablation study. The experiments were performed using the Qwen2.5-Math-7B-Instruct model on the MATH dataset. We compare our full method, EquivPruner (Finetuned w/ EM), against three variants: (1) No Pruning (vanilla MCTS baseline); (2) Pruning w/ Original Longformer (using a pre-trained Longformer-base without task-specific fine-tuning for equivalence); and (3) Pruning w/ Finetuned Longformer (w/o EM) (standard supervised fine-tuning without the EM algorithm).

The results in Figure \ref{fig:ablation} demonstrate the impact of each component. Using the Original Longformer-base for pruning (Setting 2) reduces tokens (106773 to 89998) compared to No Pruning (Setting 1), but at the cost of a accuracy drop (83.4\% to 82.4\%), indicating that a generic model is insufficient. Standard fine-tuning without EM (Setting 3) improves accuracy to 83.8\% (surpassing No Pruning) while improve token efficiency to Setting 2 (89998 to 85451), underscoring the necessity of task-specific training. Critically, our full EquivPruner method with EM-based fine-tuning (Setting 4) achieves both the highest accuracy (84.0\%) and the most significant token reduction (106773 to 74194). This highlights that both the fine-tuning process and specifically the EM algorithm are vital for maximizing EquivPruner's effectiveness in improving accuracy and token efficiency.
\section{Conclusion}
In this paper, we introduce \textit{EquivPruner}, a simple yet effective approach to address inefficient token usage in LLM reasoning search by identifying and pruning semantically equivalent actions. We also introduce MathEquiv, the first dataset specifically designed for mathematical statement equivalence, which enables the training of an effective lightweight equivalence detector. Extensive experiments demonstrate that \textit{EquivPruner} significantly reduces token consumption—for example, by 48.1\% for Qwen2.5-Math-7B-Instruct on GSM8K—while maintaining or often improving reasoning accuracy across various models and tasks. Our findings underscore the substantial benefits of managing semantic redundancy in reasoning search, offering a valuable direction for enhancing the efficiency and effectiveness of LLMs.
\section*{Limitations}
There are some limitations with our paper, which we reserve for future work.
Firstly, due to computational constraints, \textit{EquivPruner} was not evaluated on language models significantly larger than the 14B parameter scale. Secondly, our work focused on \textit{EquivPruner}'s application at inference time, and its potential integration with iterative LLM training or refinement strategies remains an area for future exploration.
\section*{Acknowledgements}
We would like to thank the anonymous reviewers for their constructive comments and valuable suggestions, which have helped us significantly improve the quality of this paper.
The work was supported by grants from the National Natural Science Foundation of China (No. U24A20253).

% 没有在更大规模的模型上测试，没有结合训练迭代方法观察效果
% 没有在常识推理、代码等任务测试，由于这些任务上的process reward model的缺失。等有PRM后补

\bibliography{custom}

\appendix
\section{Experimental Details}
\label{sec:expdetails}
\subsection{MathEquiv Dataset Generation Details}
\label{app:Dataset_Generation_Details}

The MathEquiv dataset was constructed as follows:

\textbf{Problem Sourcing and Splitting}: We selected 7,500 problems from the MATH training set \citep{hendrycks2021math}. These problems were divided into three distinct sets for EquivPruner: a training set (6,000 problems, 80\%), a validation set (750 problems, 10\%), and a test set (750 problems, 10\%). This initial split of problems ensures no data leakage between the subsequently generated step-pair datasets for EquivPruner.
\textbf{Step Pair Generation}: For each problem within these three sets, we generated candidate reasoning steps using the Qwen2.5-Math-7B-Instruct model \citep{yang2024qwen2}. This generation was performed via a Step-level Beam Search with the following parameters: beam size (k) = 8, temperature = 0.7, maximum search tree width (tree\_max\_width) = 10, maximum search tree depth (tree\_max\_depth) = 50, and maximum new tokens for generation (max\_new\_tokens) = 1024.
\textbf{Filtering}: The generated step pairs from each set were then filtered based on their Levenshtein ratio. Only pairs with a ratio between 0.75 and 0.99 (inclusive) were retained. This filtering aimed to capture meaningful variations while excluding nearly identical or overly dissimilar steps.
\textbf{Sampling and Annotation}: From the filtered pairs of each set, we randomly sampled a large number for annotation:
Training set: 80,000 pairs were annotated.
Validation set: 10,000 pairs were annotated.
Test set: 10,000 pairs were annotated. This process resulted in the final training, validation, and test sets for the MathEquiv dataset, with no overlap in the underlying problems from which the step pairs were derived.

\subsection{Evaluation Results of Pruner}
\label{eval}
We evaluated the final EquivPruner model on the held-out test set to assess the impact of incorporating the Expectation-Maximization (EM) algorithm. As shown in Table \ref{tab:em}, the experimental results confirm that integrating EM into the training process improves the pruner's performance.
\begin{table}[h]
\centering
\caption{Performance comparison of the EquivPruner on the test set, with and without the Expectation-Maximization (EM) algorithm. The best results are highlighted in bold.}
\label{tab:em}
\begin{tabular}{ccc}
\hline
Method    & Without EM & With EM          \\ \hline
F1        & 83.88\%    & \textbf{86.20\%} \\
Accuracy  & 79.40\%    & \textbf{82.97\%} \\
Precision & 84.01\%    & \textbf{85.45\%} \\
Recall    & 83.75\%    & \textbf{86.97\%} \\ \hline
\end{tabular}
\end{table}

To further assess robustness, we evaluated EquivPruner on the more challenging AIME dataset, using GPT-5.1 outputs as ground truth labels. As reported in Table~\ref{tab:aimea}, accuracy declines to 75.40\% compared to 82.97\% on MATH, while precision remains high at 91.03\%. This indicates that the pruner rarely misclassifies non-equivalent states, a property critical for avoiding performance degradation on complex problems.

\begin{table}[h]
\centering
\caption{Performance of EquivPruner on the AIME dataset.}
\label{tab:aimea}
\begin{tabular}{ccccc}
\hline
\small
Dataset & f1      & accuracy & precision & recall                                                 \\ \hline
AIME    & 77.60\% & 75.40\%  & 91.03\%   & 67.62\% \\ \hline
\end{tabular}
\end{table}
\subsection{Implementation Environment and MCTS Parameters}
\label{app:Implementation_Environment}

All experiments were conducted using PyTorch version 2.4.0. The GPU infrastructure consisted of eight NVIDIA GeForce RTX 3090 GPUs, each with 24GB, utilizing CUDA version 12.1. The central processing unit was an Intel(R) Xeon(R) Platinum 8255C CPU equipped with 96 cores.

\subsubsection{EquivPruner Training}
\label{subapp:EquivPruner_Training}

The EquivPruner model, a fine-tuned Longformer-base \citep{beltagy2020longformer}, was trained using hyperparameters selected via Bayesian optimization. The optimization aimed to maximize the `eval/f1' metric over a maximum of 10 trials. The hyperparameter search spaces are detailed in Table \ref{tab:equivpruner_hyperparams}.
\begin{table}[ht]
\centering
\caption{Hyperparameter search space for EquivPruner using Bayesian optimization.}
\label{tab:equivpruner_hyperparams}
\begin{tabular}{ll}
\hline
\textbf{Hyperparameter} & \textbf{Value or Range}  \\ \hline
Learning Rate           & {[}1e-6,5e-5{]}          \\
Batch Size              & 4                        \\
Training Epochs         & Discrete Values \{2, 3, 5\} \\
Weight Decay            & {[}0.0, 0.1{]}           \\ \hline
\end{tabular}
\end{table}

\subsubsection{MCTS Parameters}
\label{subapp:MCTS_Parameters}
The Monte Carlo Tree Search (MCTS) based evaluation hyperparameters are detailed in Table \ref{tab:mcts_hyperparams}.
These MCTS parameters (temperature, tree\_max\_width, tree\_max\_depth, simulations, PUCT values) were kept consistent across baseline and EquivPruner-enhanced evaluations unless otherwise specified.
\begin{table}[ht]
\centering
\caption{Monte Carlo Tree Search (MCTS) hyperparameters.}
\label{tab:mcts_hyperparams}
\begin{tabular}{ll}
\hline
\textbf{Hyperparameter}                                 & \textbf{Value} \\ \hline
Number of Simulations                                   & 20             \\
LLM Generation Temperature                              & 0.7            \\
LLM max\_new\_tokens         & 1024           \\
Search Tree Maximum Width                               & 10             \\
Search Tree Maximum Depth                               & 50             \\
PUCT values & 1.25           \\ \hline
\end{tabular}
\end{table}

\subsubsection{SBS Parameters}
\label{subapp:SBS_Parameters}
The Step-level Beam Search (SBS) based evaluation hyperparameters are detailed in Table \ref{tab:sbs_hyperparams}.
These SBS parameters (beam size, temperature, tree\_max\_width, tree\_max\_depth) were kept consistent across baseline and EquivPruner-enhanced evaluations unless otherwise specified.
\begin{table}[ht]
\centering
\caption{Step-level Beam Search (SBS) hyperparameters.}
\label{tab:sbs_hyperparams}
\begin{tabular}{ll}
\hline
\textbf{Hyperparameter}                                 & \textbf{Value} \\ \hline
Beam Size                                   & 3             \\
LLM Generation Temperature                              & 0.7            \\
LLM max\_new\_tokens         & 1024           \\
Search Tree Maximum Width                               & 10             \\
Search Tree Maximum Depth                               & 50             \\ \hline
\end{tabular}
\end{table}

\subsubsection{DFS Parameters}
\label{subapp:DFS_Parameters}
The Depth-First-Search (DFS) based evaluation hyperparameters are detailed in Table \ref{tab:dfs_hyperparams}.
These DFS parameters (num\_sequence, temperature, tree\_max\_width, tree\_max\_depth) were kept consistent across baseline and EquivPruner-enhanced evaluations unless otherwise specified.
\begin{table}[ht]
\centering
\caption{Depth-First-Search (DFS) hyperparameters.}
\label{tab:dfs_hyperparams}
\begin{tabular}{ll}
\hline
\textbf{Hyperparameter}                                 & \textbf{Value} \\ \hline
num\_sequence                                   & 3             \\
LLM Generation Temperature                              & 0.7            \\
LLM max\_new\_tokens         & 1024           \\
Search Tree Maximum Width                               & 10             \\
Search Tree Maximum Depth                               & 50             \\ \hline
\end{tabular}
\end{table}
\subsection{Effectiveness in Depth-First-Search}
\label{dfsapp}
To validate the versatility of our EquivPruner beyond heuristic search, we integrated it into a Depth-First Search (DFS) framework using the Qwen2.5-Math-7B-Instruct model. The results in Table \ref{tab:dfs} show that the pruner is highly effective even on systematic search algorithms.

On both MATH and GSM8K benchmarks, EquivPruner significantly reduces token consumption by over 30\% and 34.9\% respectively, while also marginally improving accuracy. This efficiency gain is achieved by preventing the redundant traversal of identical subtrees after merging equivalent states. The slight accuracy improvement suggests that by eliminating redundant paths, the search can explore the solution space more effectively, confirming that EquivPruner is a robust, algorithm-agnostic optimization.
\begin{table}[t]
\centering
\caption{Performance of EquivPruner with Depth-First-Search (DFS) using the Qwen2.5-Math-7B-Instruct model on MATH and GSM8K. EquivPruner significantly reduces token consumption (Tokens, Ratio) while improving accuracy (Acc, \%).}
\label{tab:dfs}
\begin{tabular}{cccc}
\hline
Methods                                      & Acc                  & Tokens                        & Ratio                \\ \hline
\multicolumn{1}{l}{\textit{\textbf{MATH:}}}  & \multicolumn{1}{l}{} & \multicolumn{1}{l}{\textbf{}} & \multicolumn{1}{l}{} \\
DFS                                          & 81.40                & 44229                & 100.00\%             \\
\multicolumn{1}{r}{+EquivPruner}             & 81.60                & 30943                & 69.96\%              \\ \hline
\multicolumn{1}{l}{\textit{\textbf{GSM8K:}}} & \multicolumn{1}{l}{} & \multicolumn{1}{l}{}          & \multicolumn{1}{l}{} \\
DFS                                          & 95.60                & 14821                         & 100.00\%             \\
\multicolumn{1}{r}{+EquivPruner}             & 95.68                & 9648                          & 65.10\%              \\ \hline
\end{tabular}
\end{table}
\subsection{Experiments on AIME 2024 dataset}
\label{app:aime}
We conducted additional experiments on the AIME dataset using both the Mistral-7B-sft and Qwen2.5-Math-7B-Instruct models. As shown in Table \ref{tab:aime}, these results on AIME consistently show that our EquivPruner effectively reduces token consumption across different models and search algorithms without compromising performance.
\begin{table}[ht]
\centering
\caption{Performance of EquivPruner with MCTS using the Qwen2.5-Math-7B-Instruct model and Mistral-7B-sft model on AIME 2024.}
\label{tab:aime}
\begin{tabular}{cccc}
\hline
Methods                                      & Acc   & Tokens     \\ \hline
\multicolumn{1}{l}{\textit{\textbf{Qwen2.5-Math-7B-Instruct:}}}  &       &        \\
Vanilla MCTS                       & 16.67 & 220446   \\
\multicolumn{1}{r}{+ EquivPruner}            & 16.67 & 204654    \\ \hline
\multicolumn{1}{l}{\textit{\textbf{Mistral-7B-sft:}}} &       &        \\
Vanilla MCTS                       & 0.00 & 87002   \\
\multicolumn{1}{r}{+ EquivPruner}            & 3.33 & 68832     \\ \hline
\end{tabular}
\end{table}
\subsection{Failure Mode Analysis}
\label{fma}
To better understand the conditions under which EquivPruner is most effective, we conducted a correlation analysis on 100 randomly sampled problems from the MATH dataset. Table~\ref{tab:correlation} reports Pearson correlation coefficients among key metrics, including pruning rate, token saving ratio, tree depth, and token consumption.

Pruning rate exhibits a moderate negative correlation with token consumption in failure cases ($r = -0.363$, $p < 0.001$) and with tree depth in failure cases ($r = -0.377$, $p < 0.001$). This indicates that when pruning fails, it tends to occur on problems characterized by shallower search trees and lower token usage. Conversely, pruning rate shows a weak negative correlation with problem difficulty as measured by the number of reasoning levels ($r = -0.207$, $p = 0.039$), and a moderate positive correlation with token saving ratio ($r = 0.396$, $p < 0.001$), confirming that higher pruning rates translate to greater efficiency gains.

A qualitative examination of false positives further revealed that in all four identified failure cases, the shallower trees nonetheless contained nodes with substantially longer sequences, challenging the equivalence classifier. These findings collectively suggest that pruning benefits are most pronounced for problems with inherently deeper search trees, and that future improvements may be achieved by augmenting training data with longer-sequence examples.

\begin{table*}[h]
\centering
\caption{Pearson correlations among pruning metrics on MATH dataset ($n=100$).}
\label{tab:correlation}
\begin{tabular}{lcc}
\hline
Variable Pair & Pearson $r$ & $p$-value \\
\hline
Pruning rate \& Token saving ratio & 0.396 & $<0.001$ \\
Pruning rate \& Token consumption (failure cases) & $-0.363$ & $<0.001$ \\
Pruning rate \& Tree depth (failure cases) & $-0.377$ & $<0.001$ \\
Pruning rate \& Problem levels & $-0.207$ & $0.039$ \\
Token saving ratio \& Tree depth (failure cases) & $-0.115$ & $0.253$ \\
\hline
\end{tabular}
\end{table*}
\subsection{Analysis of Formula Extraction-based Pruning}
\label{app:Formula_Extraction}
\begin{table}[]
\centering
\caption{Performance comparison of the formula extraction-based pruning method and EquivPruner on the GSM8K dataset (Model: Qwen2.5-Math-7B-Instruct).}
\label{tab:Formula_Extraction}
\begin{tabular}{rcc}
\hline
\multicolumn{1}{c}{Methods}                                                          & Acc   & Tokens \\ \hline
\multicolumn{1}{c}{Vanilla MCTS}                                                     & 96.44 & 34826  \\
\begin{tabular}[c]{@{}r@{}}+ Formula Extraction \\ with String Match\end{tabular}    & 96.44 & 25632  \\
\begin{tabular}[c]{@{}r@{}}+ Formula Extraction \\ with Embedding Model\end{tabular} & 96.44 & 25336  \\
+EquivPruner (Ours)                                                                  & 96.59 & 18071  \\ \hline
\end{tabular}
\end{table}
An intuitive pruning approach in mathematical reasoning involves extracting arithmetic equations from the generated text and directly checking for equivalence. We investigated this formula extraction-based method to evaluate its feasibility and performance.

A primary challenge of this approach is its dependency on the language model's output format. The style of arithmetic equations can be highly diverse across models, making consistent extraction via simple rules difficult. However, we observed that for simpler problems, such as those in the GSM8K dataset, the Qwen2.5-Math-7B-Instruct model often wraps formulas in consistent delimiters . This formatting consistency enabled a targeted experiment.
In our experiment, we first extracted formulas from the generated solutions and then applied two techniques to check for equivalence: (1) direct string matching of the extracted formulas and (2) comparing the formulas using an embedding model (MATH-Similarity).

As show in Table \ref{tab:Formula_Extraction}, the results indicate that both formula-based pruning methods successfully reduced the token count compared to Vanilla MCTS—by 26.4\% for string matching and 27.2\% for the embedding model. However, neither method yielded an improvement in accuracy, achieving the same 96.44\% as the vanilla baseline. In contrast, our EquivPruner achieved a significantly greater token reduction of 48.1\% while also increasing accuracy to 96.59\%.
This experiment highlights the limitations of the formula extraction approach. Even when formatting is consistent enough for reliable extraction, it fails to outperform EquivPruner in either efficiency or effectiveness. Its reliance on specific delimiters makes the method brittle and not generalizable to other models or more complex problem types where output may be less structured. This justifies our focus on the more robust and versatile execution-based approach implemented in EquivPruner.
\begin{figure*}[!htbp]
\centering
\scalebox{1}{
\begin{tcolorbox}
Please determine whether the following two sentences are semanticly equivalent, and return 0: Not equivalent at all; 1: May not be equivalent; 2: Can't judge; 3: May be equivalent; 4: Exactly equivalent.
Please explain the reason, reflect, and provide a more accurate result.
Please output in the following Python dictionary format:\\
\{
    "reasoning\_step": "The reasoning process of the model",
    "result": "Final result" (int)
\}\\
Question:\\
\textbf{Sentence1: }
The inner sum is a geometric series with the first term $\frac{{1}}{{2^{{k^2-k+1}}}}$ and common ratio $\frac{{1}}{{2}}$, and it has $2k$ terms. The sum of a geometric series is given by:
$$\sum_{{n=a}}^{{a+b-1}} ar^n = a \frac{{1-r^b}}{{1-r}}$$

\textbf{Sentence2: }
The inner sum $\sum_{{n=k^2-k+1}}^{{k^2+k}} \frac{{1}}{{2^n}}$ is a geometric series with the first term $\frac{{1}}{{2^{{k^2-k+1}}}}$ and common ratio $\frac{{1}}{{2}}$, and it has $2k$ terms. The sum of a geometric series is given by
$$\sum_{{n=a}}^{{a+r-1}} ar^n = a \frac{{1-r^r}}{{1-r}}$$

Output:\\
\{
    "reasoning\_step": "Sentence1 uses the formula $\sum_{{n=a}}^{{a+b-1}} ar^n = a \frac{{1-r^b}}{{1-r}}$, while Sentence2 uses $\sum_{{n=a}}^{{a+r-1}} ar^n = a \frac{{1-r^r}}{{1-r}}$. The variables and the structure of the formulas are different, which affects the semantic equivalence. Additionally, the limits of the summation in Sentence2 are explicitly given as $\sum_{{n=k^2-k+1}}^{{k^2+k}} \frac{{1}}{{2^n}}$, which is not present in Sentence1. These differences indicate that the sentences are not semantically equivalent.",
    "result": 0
\}\\
Question:\\
\textbf{Sentence1: }
The amplitude is the distance from the midline to the maximum value (or the minimum value). Therefore, we can calculate the amplitude as follows:
$${{Amplitude}} = {{Maximum value}} - {{Midline}} = 3 - 1 = 2$$
Since the amplitude is $|a|$, we have:|a| = 2. \\ Since $a$ is a positive constant, we conclude:a = 2

\textbf{Sentence2: }
The amplitude of the sine wave is half the distance between the maximum value and the minimum value. Therefore, we can calculate the amplitude as follows:
$${{Amplitude}} = \frac{{{{Maximum value}} - {{Minimum value}}}}{{2}} = \frac{{3 - (-1)}}{{2}} = \frac{{3 + 1}}{{2}} = \frac{{4}}{{2}} = 2$$ Since the amplitude is $|a|$, we have:|a| = 2. Since $a$ is a positive constant, we conclude: a = 2

Output:
\{
    "reasoning\_step": "Sentence1 defines the amplitude as the distance from the midline to the maximum value, calculating it as ${{Amplitude}} = {{Maximum value}} - {{Midline}} = 3 - 1 = 2$. Sentence2 defines the amplitude as half the distance between the maximum and minimum values, calculating it as ${{Amplitude}} = \frac{{{{Maximum value}} - {{Minimum value}}}}{{2}} = \frac{{3 - (-1)}}{{2}} = 2$. While both methods yield the same result ($a = 2$), the definitions and calculations are fundamentally different. This difference in methodology means the sentences are not semantically equivalent.",
    "result": 0
\}\\
Question:\\
\textbf{Sentence1:} 
\{sentence1\}\\
\textbf{Sentence2:}
\{sentence2\}

Output:
\end{tcolorbox}
}
\caption{Complete prompt for labeling.}
\label{prompt}
\end{figure*}

\begin{figure*}[!htbp]
\centering
\scalebox{1}{
\begin{tcolorbox}
\textbf{Sentence1:} 
Thus, the sum of the elements in the $ n $-th row of Pascal's triangle is $ 2^n $. The function $ f(n) $ is the base-10 logarithm of this sum, so we have:\\
\textbf{Sentence2:}
This means that the sum of the elements in the $ n $-th row of Pascal's triangle is $ 2^n $. Therefore, $ f(n) $ is the base-10 logarithm of $ 2^n $:\\
\textbf{result:} 4\\
\textbf{reasoning\_step:} Both sentences convey the same mathematical property of Pascal's triangle, specifically that the sum of the elements in the $ n $-th row is $ 2^n $. Sentence1 states this as a conclusion, while Sentence2 presents it as a clarification. They both express the same final idea regarding the function $ f(n) $ being the base-10 logarithm of this sum. Therefore, they are semantically equivalent.
\end{tcolorbox}
}
\caption{An Exactly Equivalent Example on MathEquiv Dataset.}
\label{4example}
\end{figure*}
\begin{figure*}[!htbp]
\centering
\scalebox{1}{
\begin{tcolorbox}
\textbf{Sentence1:} 
Since $ u \geq \sqrt{3} $ and $ 2 \geq \sqrt{3} $, $ u = 2 $ is a valid critical point. We evaluate $ g(u) $ at $ u = 2 $:\\
\textbf{Sentence2:}
Since $ u \geq \sqrt{3} $ and $ 2 \geq \sqrt{3} $, $ u = 2 $ is a valid critical point. We evaluate $ g(u) $ at this critical point and at the boundary $ u = \sqrt{3} $:\\
\textbf{result:} 0\\
\textbf{reasoning\_step:} Sentence1 explicitly evaluates $ g(u) $ at $ u = 2 $ only, while Sentence2 mentions that $ g(u) $ is evaluated at both the critical point (which is also $ u = 2 $) and at the boundary $ u = \sqrt{3} $. The additional evaluation at the boundary in Sentence2 introduces a difference that affects the overall semantic message, making them not semantically equivalent.
\end{tcolorbox}
}
\caption{A Not Equivalent Example on MathEquiv Dataset.}
\label{1example}
\end{figure*}
\end{document}